\documentclass[conference]{IEEEtran}
\IEEEoverridecommandlockouts
\usepackage{cite}
\usepackage{amsmath,amssymb,amsfonts}
\usepackage{url}
\usepackage{graphicx}
\usepackage{textcomp}
\usepackage{graphicx} 
\usepackage{epsfig} 
\usepackage{amsmath} 
\usepackage{float}      
\usepackage{subcaption}
\usepackage{caption}
\usepackage{newfloat}
\DeclareFloatingEnvironment[name={Supplementary Figure}]{suppfigure}
\def\BibTeX{{\rm B\kern-.05em{\sc i\kern-.025em b}\kern-.08em
    T\kern-.1667em\lower.7ex\hbox{E}\kern-.125emX}}
\begin{document}

\title{An Energy Balance Based Method for Parameter Identification of a Free-Flying Robot Grasping An Unknown Object\\
\thanks{*This work was supported
	by the FCT project [UID/EEA/50009/2013]}
}

\author{\IEEEauthorblockN{Monica Ekal, Rodrigo Ventura}
	\IEEEauthorblockA{\textit{Institute for Systems and Robotics} \\
		\textit{Instituto Superior Tecnico}\\
		Lisbon, Portugal \\
		\{mekal, rodrigo.ventura\}@isr.tecnico.ulisboa.pt}

}

\maketitle


\begin{abstract}
	The estimation of inertial parameters of a robotic system is crucial for better trajectory tracking performance, specially when model-based controllers are used for carrying out precise tasks. In this paper, we consider the scenario of grasping an object of unknown properties by a free-flyer space robot with limited actuation. The problem is to find the inertial parameters of the complete system after grasping has been performed. Excitation is provided in inertial space, and the excitation trajectories  are found by optimization. Truncated Fourier series are used to represent the reference as well as tracked trajectory. An approach based on the energy balance between the actuation work and the rate of change of kinetic energy is introduced to calculate the number of harmonics in the Fourier series used to represent the executed trajectory, while trying to find a balance between accounting for saturation effects and keeping out noise. The effect of input saturation on parameter estimation is also studied. Simulation results using the Space CoBot free-flyer robot are presented to show the feasibility of the approach.
\end{abstract}

\section{INTRODUCTION}
Robotic On-Orbit Servicing (OOS) missions like Orbital Express\cite{c0} and ETS-VII\cite{c001} consist of free-flyer robots entrusted with challenging tasks such as maintenance, removal of debris and docking. These tasks are precise and also due to the coupling that exists between the dynamics of the robot manipulator and its base, they demand high-level controllers based on the model of the system. Therefore, it becomes important to have knowledge of the inertial parameters like mass, inertia and centre of mass of such free-flying robotic systems.

The problem of parameter identification is not new. In literature, methods based the Newton-Euler equations of motion have been used to tackle this problem in the case of terrestrial fixed manipulators, \cite{c1}, \cite{c2}, \cite{c7} and quadrotors \cite{c3} among others. This method has been extended in case of space manipulator systems in works such as \cite{a8}, where the virtual manipulator approach has been used for modelling the system. Another approach is to use the property of conservation of momentum. Ma et al. \cite{a3} have used angular momentum conservation to estimate inertial parameters of the spacecraft only, with knowledge of the payload and arm parameters.  In works such as \cite{c6}, excitation is provided only in the joints of the arms to identify complete parameters of a manipulator system in the free-floating mode. However, they have used momentum control devices to maintain non-zero angular momentum for free-floating systems. In \cite{a1}, Kazuya et al. have used the effect of the gravity gradient torque to complement parameter estimation based on  momentum conservation in free-flying mode.  Rackl et al. \cite{a5} as well as Morutso et al. \cite{a4} have compared the performance of existing methods based on the two above-mentioned approaches.
Thus, when it comes to space systems, the use of the property of momentum conservation is more common. This is because the approach does not require measurements of accelerations of the robot base and manipulator torques, which contain noise and are difficult to measure, respectively. Magnitude and direction of the applied thrust would also be needed, if the robot is in free-flying mode.  However, the advantage of the equations-of-motion method is that the equations can be expressed linearly in terms of the dynamic parameters, which makes a linear least squares solution possible. 
\begin{figure}
	\centering
	\includegraphics[width=80mm,scale=0.5]{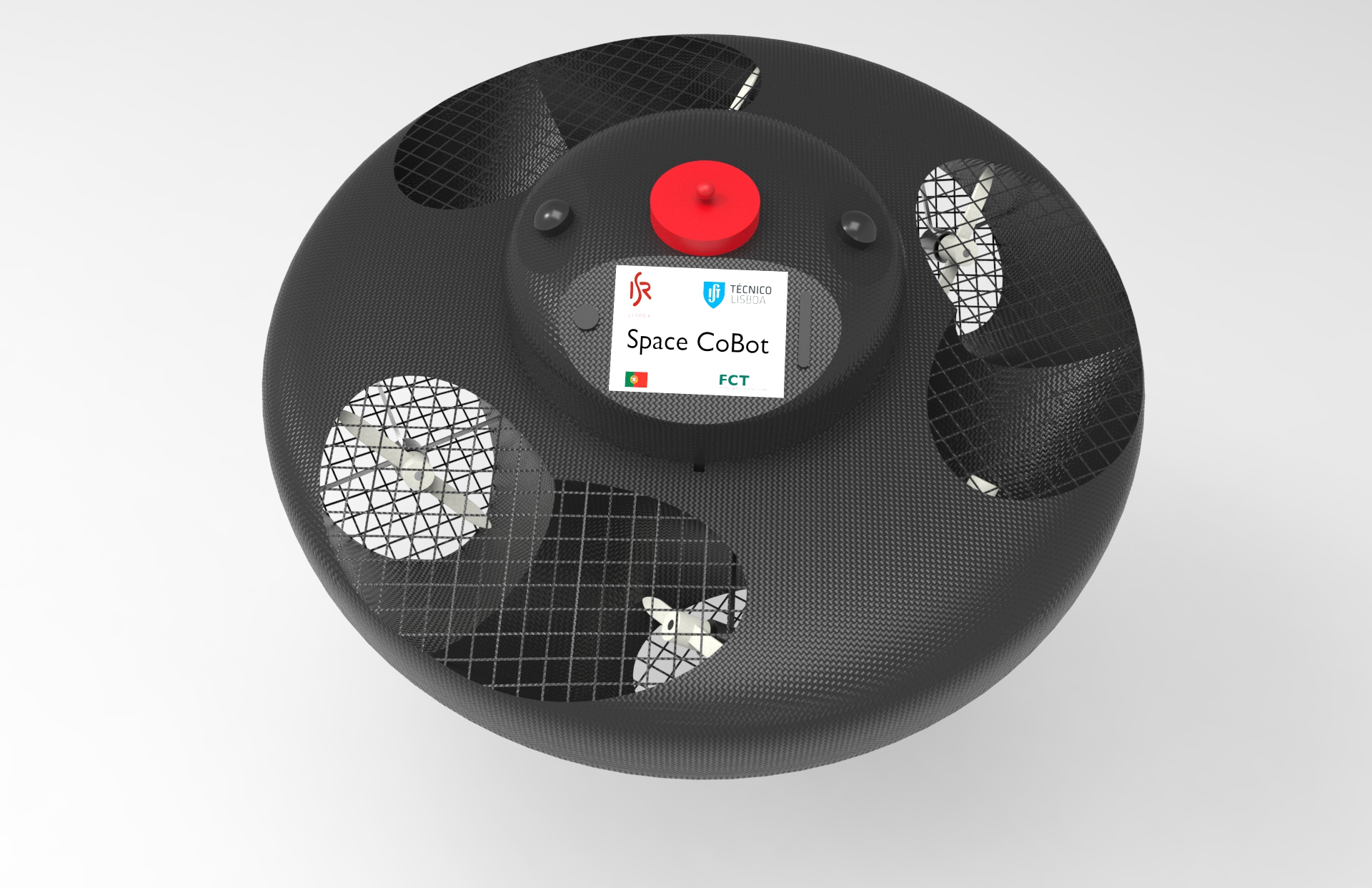}
	\caption{A free-flying robot: The Space CoBot}\label{pic1}
	\vspace{-1.8em}	
\end{figure}

In previous research, methods are proposed to estimate a maximum number of parameters of the robot base and the manipulator. Often, external forces and torques are set to zero by considering free-floating mode. The goal of this paper is to estimate inertial parameters - mass, centre of mass and inertia tensor - of the robot-object body, after the robot has grasped an unknown object. As a first step, the unmodeled load is considered to be attached directly to the base. Our method extends \cite{c1}, where it was used for parameter identification of ground-based manipulators. The equations of motion approach is used, and the executed trajectory is represented using Fourier series so that the measured velocity and acceleration can be found by analytical differentiation. We use more harmonics for estimating the tracked excitation trajectory as compared to the reference, and a method to select the number of harmonics based on the difference between the kinetic energy rate and power is proposed. This method of parameter estimation is specially relevant for free-flying robots in environments present in the interior of orbiting stations, like the NASA projects SPHERES \cite{c01} and Astrobee \cite{c02}, which are intended to function inside the International Space Station (ISS). In this paper, we consider the example of the Space CoBot \cite{c16} to evaluate the proposed method.

The paper is structured as follows:
Dynamics of the robot are detailed in Section \ref{sec:2}. In Section \ref{sec:3}, the parameter estimation problem is formulated. Generation of the exciting trajectories using optimization criteria and the use of Model Predictive control for tracking them are described in Section \ref{sec:4}. Section \ref{sec:5} discusses the steps involved in estimating the parameters from obtained sensor data. In Section \ref{sec:6}, we present results of our simulations to show feasibility of the approach,
while Section \ref{Conclusion} provides concluding remarks about it and plans for future work.

\section{DYNAMICS OF THE FREE-FLYING SYSTEM}\label{sec:2}

\subsection{The Space CoBot}

The Space CoBot is a free-flying robot, designed with the goal of assisting astronauts in indoor environments, like inside an orbiting space station. It works on electric propulsion, with 6 motors arranged to provide holonomic motion. Placement of the motors is decided using multi-criteria optimization, such that the range of forces and torques in all directions are maximized \cite{c5}.  It would be used to carry out chores and maintenance on the space station, such as tools and components handling, remote inspection and debris scavenging \cite{c16}. This makes it important for the robot to be capable of autonomous grasping and mobile manipulation. 
\subsection{Rigid-body dynamics} \label{dyn}
For the sake of simplicity, only the dynamics of the robot have been modeled here. We consider that an unmodeled load is attached to the robot. This problem can be extended further to include dynamics of the manipulator.
\begin{figure*}
	\centering
	\minipage{0.33\linewidth}
	\includegraphics[width=\linewidth]{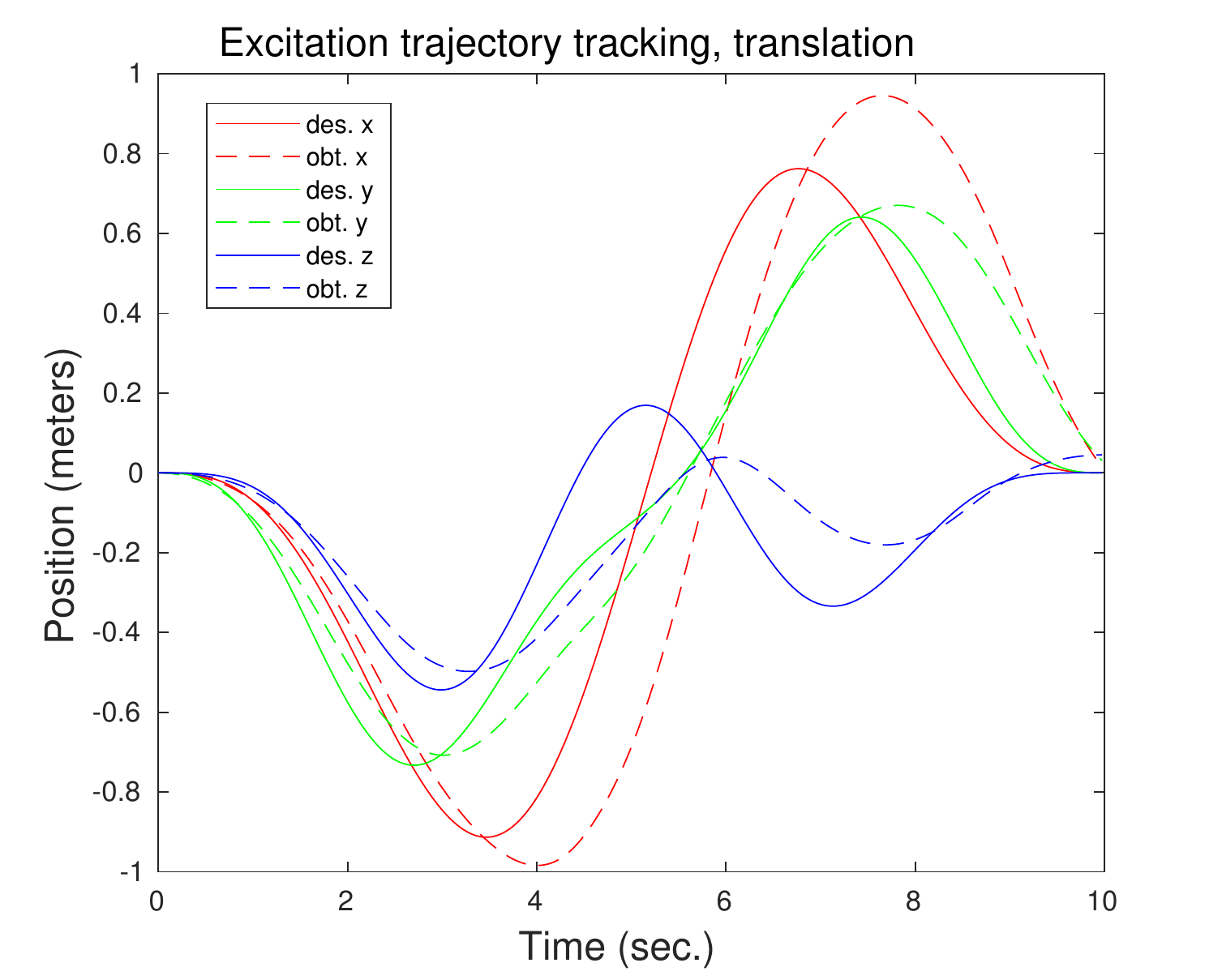}
	\endminipage%
	\minipage{0.33\linewidth}
	\includegraphics[width=\linewidth]{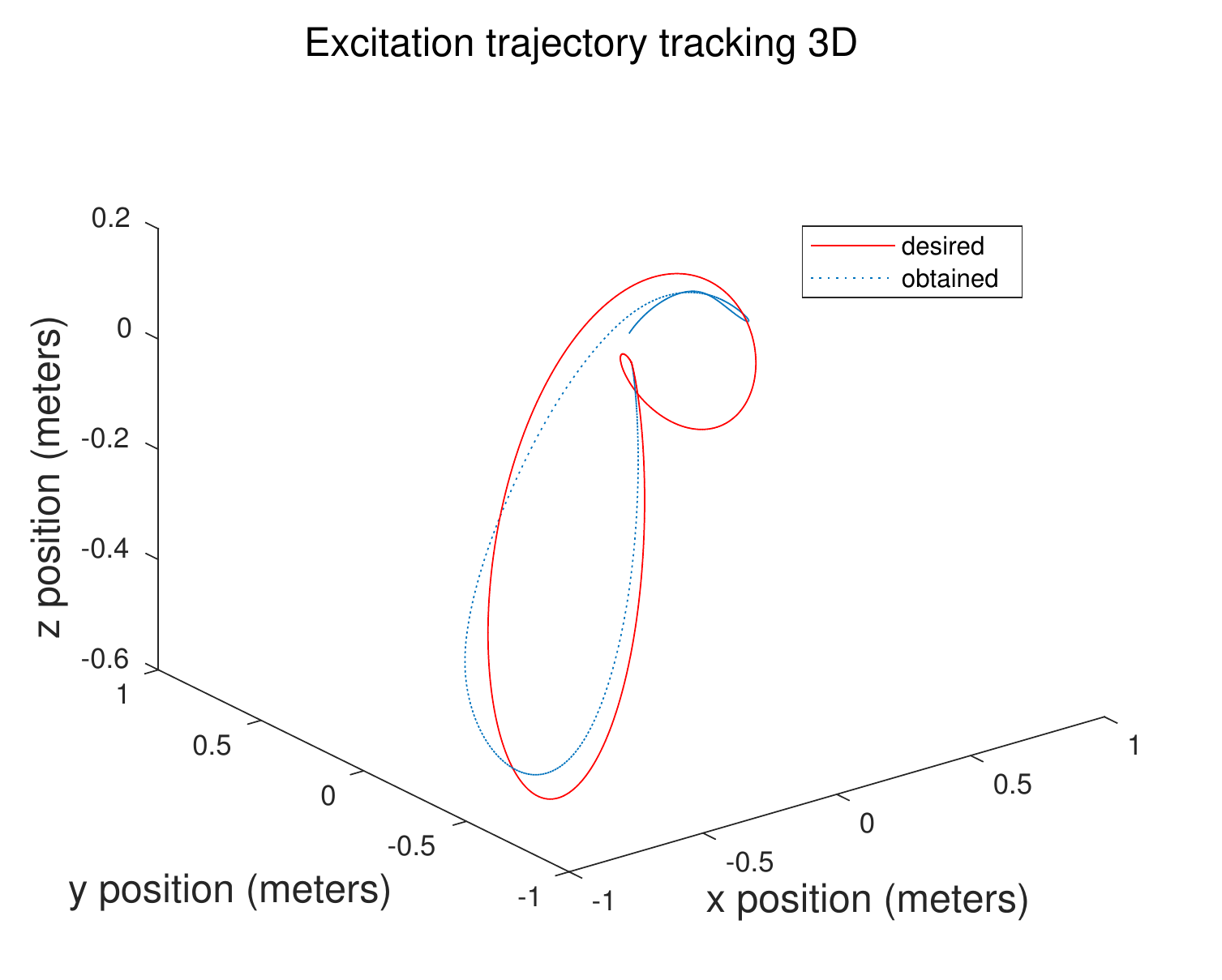}
	\endminipage%
	\minipage{0.33\linewidth}
	\includegraphics[width=\linewidth]{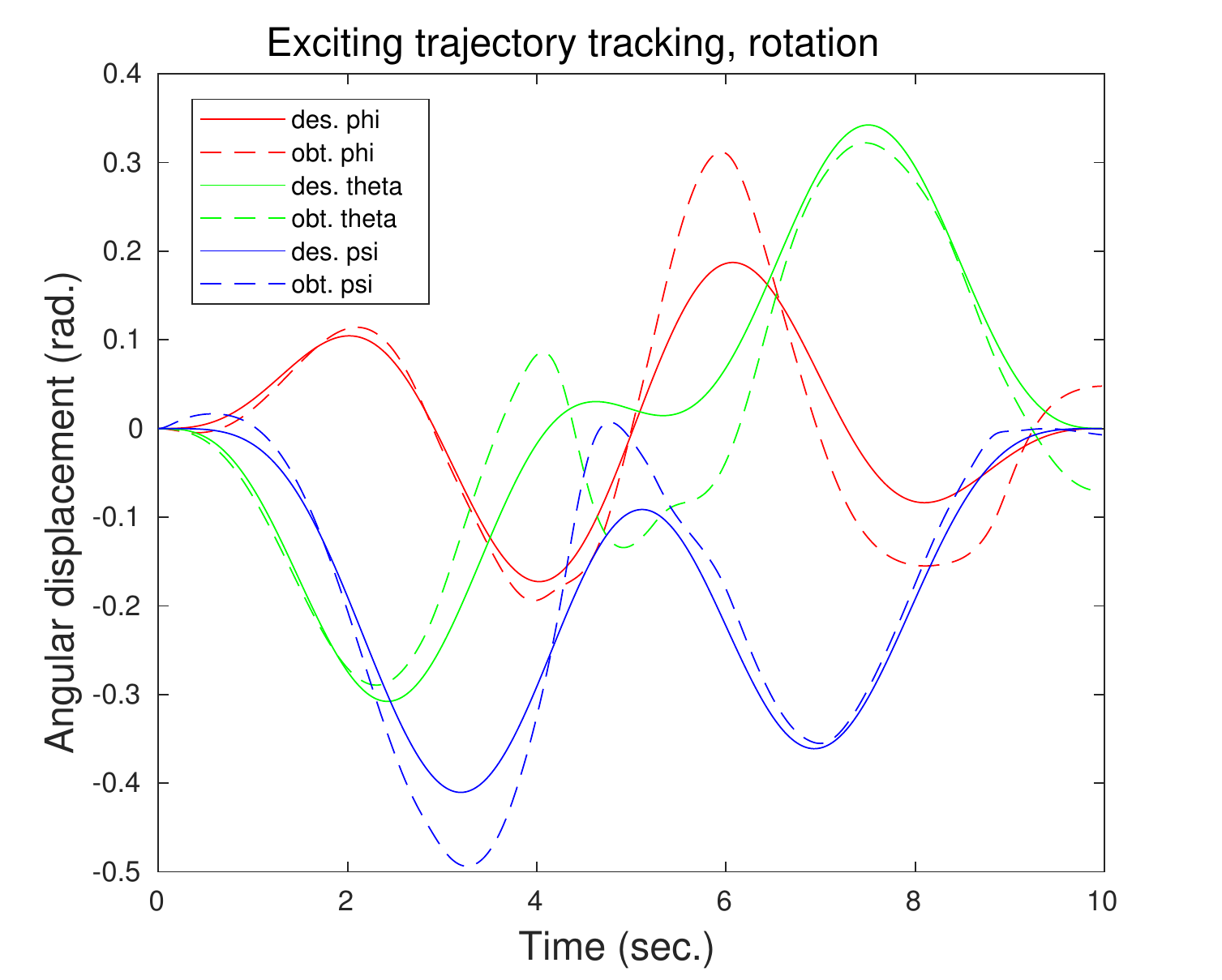}
	\endminipage
	\caption{The desired Excitation trajectory and the one tracked by the robot }\label{fig:2}
	
\end{figure*}
\\
\indent Vectors of generated force $\textbf{F} $ and moments $\textbf{M} $ for the robot are given as
\begin{equation}\label{eq01}
\left( \begin{array}{c}
\textbf{{F}} \\
\textbf{{M}} 
\end{array}
\right) = \mathbf{A} \textbf{\textit{u}}
\end{equation}
where $\mathbf{A}$ is the 6x6 actuation matrix, also known as the mixing matrix, composed of the contributions of each propeller to the net forces and moments, and $\textbf{\textit{u}} = [u_1~ ...~ u_6]^T $ is the vector of actuation inputs for each motor \cite{c5}.
The body frame, $\mathcal{B}$, of the CoBot is situated at its center of mass, $\textrm{P}_{c}$. $\textrm{P}_{s}$ is the changed center of mass of the system after the grasping action has been performed. Vectors $\textbf{\textit{p}}_{c}$ and $\textbf{\textit{p}}_{s}$, are the position vectors of these two centers of mass, with respect to the inertial frame $\mathcal{I}$. They are related to the vector denoting the offset between them, $\textbf{\textit{p}}_{off}$, which is expressed in the body frame, $\mathcal{B}$ as:
\begin{equation}\label{eq0}
\textbf{\textit{p}}_{s} = \textbf{\textit{p}}_{c} + \mathbf{R} \textbf{\textit{p}}_{off} 
\end{equation}
\begin{equation}\label{eq00}
\dot{\textbf{\textit{p}}}_{s} = \dot{\textbf{\textit{p}}_{c}} + \mathbf{R} \left(\pmb{\omega}\times\textbf{\textit{p}}_{off}\right)
\end{equation}
where $\mathbf{R}$ is the rotation matrix of frame $\mathcal{B}$ with respect to frame $\mathcal{I}$.
\newline Newton-Euler equations for acceleration of the vehicle's center of mass $\textrm{P}_{c}$ with respect to the inertial frame and the rotation of system in the body frame are:
\begin{equation}\label{eq1}
m\left\{\ddot{\textbf{\textit{p}}}_{c} + \mathbf{R}\Bigg(\dot{\pmb{\omega}}\times\textbf{\textit{p}}_{off} + \Big(\pmb{\omega}\times\left(\pmb{\omega}\times\textbf{\textit{p}}_{off}\right) \Big) \Bigg) \right\} = \mathbf{R}\textbf{F}
\end{equation}
\begin{equation}\label{eq2}
\mathbf{J}_s\dot{\pmb{\omega}}  + \pmb{\omega}\times\mathbf{J}_s\pmb
{\omega} + \textbf{\textit{p}}_{off}\times\mathbf{F} = \mathbf{M}
\end{equation}
where $m$ stands for mass of the system, $\mathbf{J}$ for its moment of inertia and the operator $\times$ denotes cross product. $\pmb{\omega}$ is the angular velocity of the vehicle expressed in the body frame $\mathcal{B}$. 

\section{ESTIMATION OF PARAMETERS}\label{sec:3}
In this section, we describe the method for estimation of inertial parameters using measured data.
As done in \cite{c2}, parallel axis theorem can be used to express $\mathbf{J}$ about the origin of body frame $P_{c}$ instead of the robot-object system's centre of mass $P_{s}$ as:
\begin{equation}\label{eq3}
\mathbf{J}_c = \mathbf{J}_s + m[ (\textbf{\textit{p}}_{off}^T\textbf{\textit{p}}_{off})I - (\textbf{\textit{p}}_{off } \textbf{\textit{p}}_{off }^T)] 
\end{equation} 
Substituting (\ref{eq1}) in (\ref{eq2}) and using (\ref{eq3}), (\ref{eq2}) can be written as:
\begin{equation}
\mathbf{J}_c\dot{\pmb{\omega}}  + \pmb{\omega}\times\mathbf{J}_c\pmb{\omega} + m\textbf{\textit{p}}_{off}\times\mathbf{R}^{-1}\textbf{\textit{p}}_c = \mathbf{M}
\end{equation}

The Newton-Euler equations thus become linear in terms of the inertial parameters to be estimated. They can be written in a compact form as:
\begin{equation}
\pmb{\gamma}(\mathbf{X},\dot{\mathbf{X}},\ddot{\mathbf{X}})\pmb{\pi} = \pmb{\tau}
\end{equation}
where $\pmb{\gamma}(\mathbf{X},\dot{\mathbf{X}},\ddot{\mathbf{X}})$ is the regressor  matrix with $\mathbf{X} = [x,y,z,\phi,\theta,\psi]$, and $\psi,\theta,\phi$ being the Z-Y-X Euler Angles used to express orientation relative to frame $\mathcal{I}$.
\begin{equation}
\pmb{\gamma} = 
\left[ \begin{array}{ccc}
\mathbf{R}^{-1}\ddot{\textbf{\textit{p}}}_c & \mathbf{S}(\dot{\pmb{\omega}}) + \mathbf{S}(\pmb{\omega})\mathbf{S}(\pmb{\omega}) &  \mathbf{0}_{3\times 6}\\
\mathbf{0}_{3\times 1} & -\mathbf{S}(\mathbf{R}^{-1}\ddot{\textbf{\textit{p}}}_c) & [*\dot{\pmb{\omega} }]+\mathbf{S}(\pmb{\omega})[*{\pmb{\omega }}]
\end{array}
\right]\end{equation}
Here, $\mathbf{S}(\pmb{\omega})$ represents the skew-symmetric matrix,
\begin{equation}
\mathbf{S}(\pmb{\omega}) = \left[\begin{array}{ccc}
0 & -\omega_z &\omega_y\\ \omega_z & 0 &-\omega_x \\ -\omega_y &\omega_x & 0
\end{array}\right]
\end{equation}
and $[*\pmb{\omega}]\mathit{J}$ is obtained by calculating $\mathbf{J}_c\pmb{\omega}$ and re-arranging the product as:
\begin{equation}\label{eq}
\mathbf{J}_c\pmb{\omega} = \left[\begin{array}{cccccc}
\omega_x & \omega_y & \omega_z & 0 & 0& 0\\
0 & \omega_x & 0 & \omega_y & \omega_z &0 \\
0 & 0 & \omega_x & 0 & \omega_y & \omega_z
\end{array}\right] \left[\begin{array}{c}
J_{xx} \\ J_{xy} \\ J_{xz}\\ J_{yy}\\ J_{yz}\\ J_{zz}
\end{array}\right]
\end{equation}
$[*\pmb{\omega}]$ is the first matrix from (\ref{eq}). $\pmb{\pi}$ is the 10x1 vector of parameters to be estimated:
\begin{equation}
\pmb{\pi} = 
\left[\begin{array}{cccccccccc}
m &m\mathbf{p}_{off}^T & {J}_{xx}& {J}_{xy} &{J}_{xz}& {J}_{yy}& {J}_{yz}& {J}_{zz}
\end{array}\right]^T
\end{equation}
and $\pmb{\tau}$ is the vector of the applied forces and moments
\begin{equation}
\pmb{\tau} = \left[\begin{array}{c}
\mathbf{F} \\ \mathbf{M}
\end{array}\right]
\end{equation}

When $N$ measurements of the states ($x, y, z$ and Euler angles, $\phi, \theta, \psi$) are available, a least square problem can be set up as
\begin{equation}\label{eq5}
\mathbf{W}\pmb{\pi} = \mathbf{b}
\end{equation}
where
\begin{equation}\label{eq14}
\mathbf{W} = 
\left[
\begin{array}{c}
\pmb{\gamma}(X(1),\dot{X}(1),\ddot{X}(1))\\ \vdots \\ \pmb{\gamma}(X(N),\dot{X}(N),\ddot{X}(N))
\end{array}
\right]
\end{equation}
and 
\begin{equation}\label{eq15}
\mathbf{b} = 
\left[ \begin{array}{c}
\tau(1)\\ \vdots \\\tau(N)
\end{array}
\right]
\end{equation}
Then the solution vector $\hat{\pmb{\pi}}$ is given by 
\begin{equation}\label{eq16}
\hat{\pmb{\pi}} = (\mathbf{W}^T\mathbf{W})^{-1}\mathbf{W}^T\mathbf{b}
\end{equation}
where the regressor matrix, $\mathbf{W}$, must be full-ranked. This means that the result of the parameter identification depends on $\mathbf{W}$, which is a function of $X,\dot{X},\ddot{X}$. Moreover, estimating velocity and acceleration from real data is non-trivial due to sensor noise. Thus, appropriate excitation and measured data processing is needed to obtain a good estimate.

\section{EXCITATION TRAJECTORY}\label{sec:4}
In this section, we summarize the procedure for calculation and tracking of the excitation trajectory.
\subsection{Generation of the excitation trajectory} \label{ssec:exc}
As in \cite{c1}, the approach for calculating excitation trajectories is based on Fourier series. Benefits of using Fourier Series are that the resulting trajectory is band-limited and periodic. This makes it possible to average, over multiple periods, the data collected from the sensors, resulting in a better signal-to-noise ratio. Further, specifying the bandwidth aids in identifying the spectral lines that contain measurement information in the frequency domain, at the time of noise filtering.
\begin{figure*}
	
	\centering
	\minipage{0.333\linewidth}
	\includegraphics[width=\linewidth]{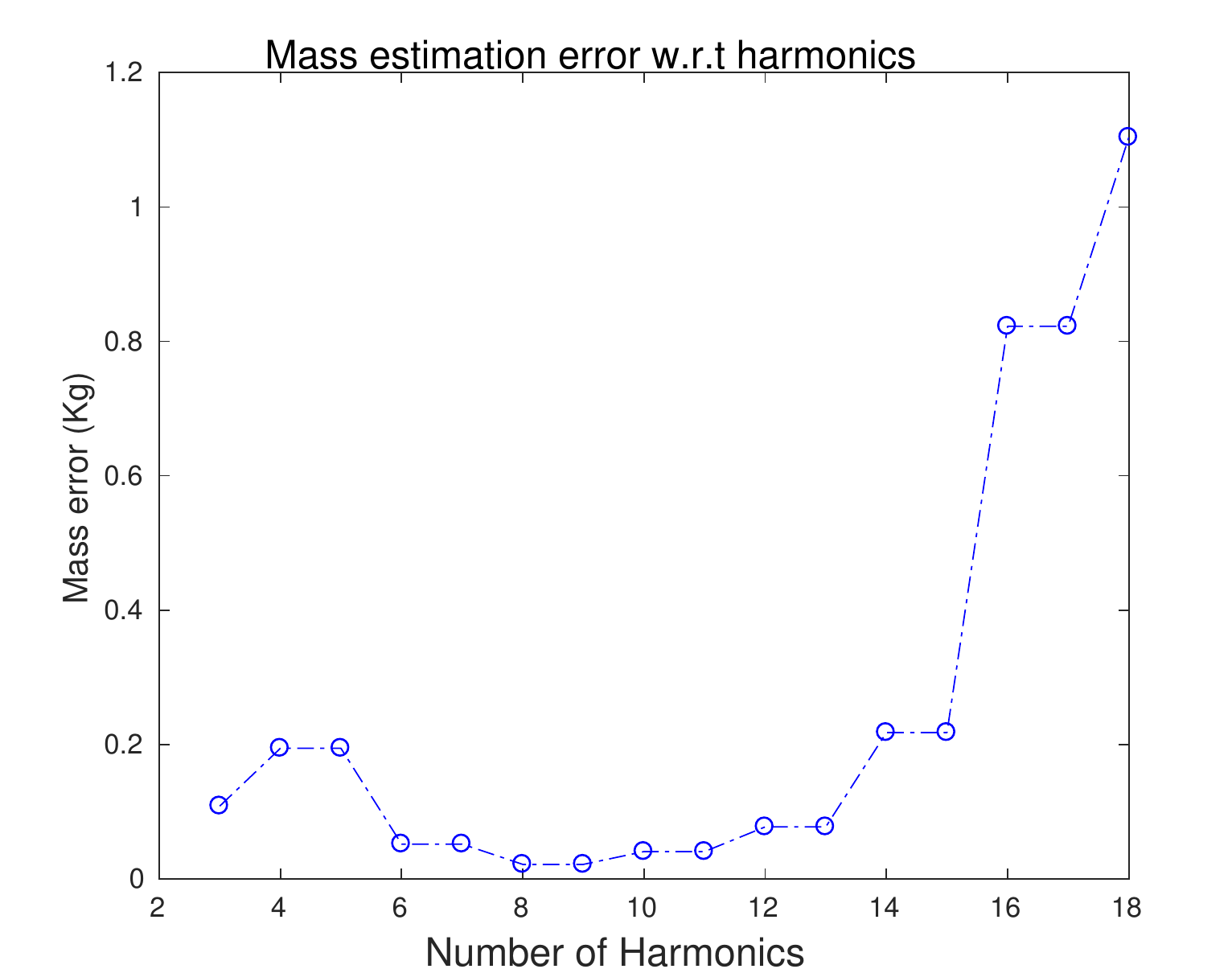}
	\subcaption{}\label{ME} 
	\endminipage%
	\minipage{0.333\linewidth}
	\includegraphics[width=\linewidth]{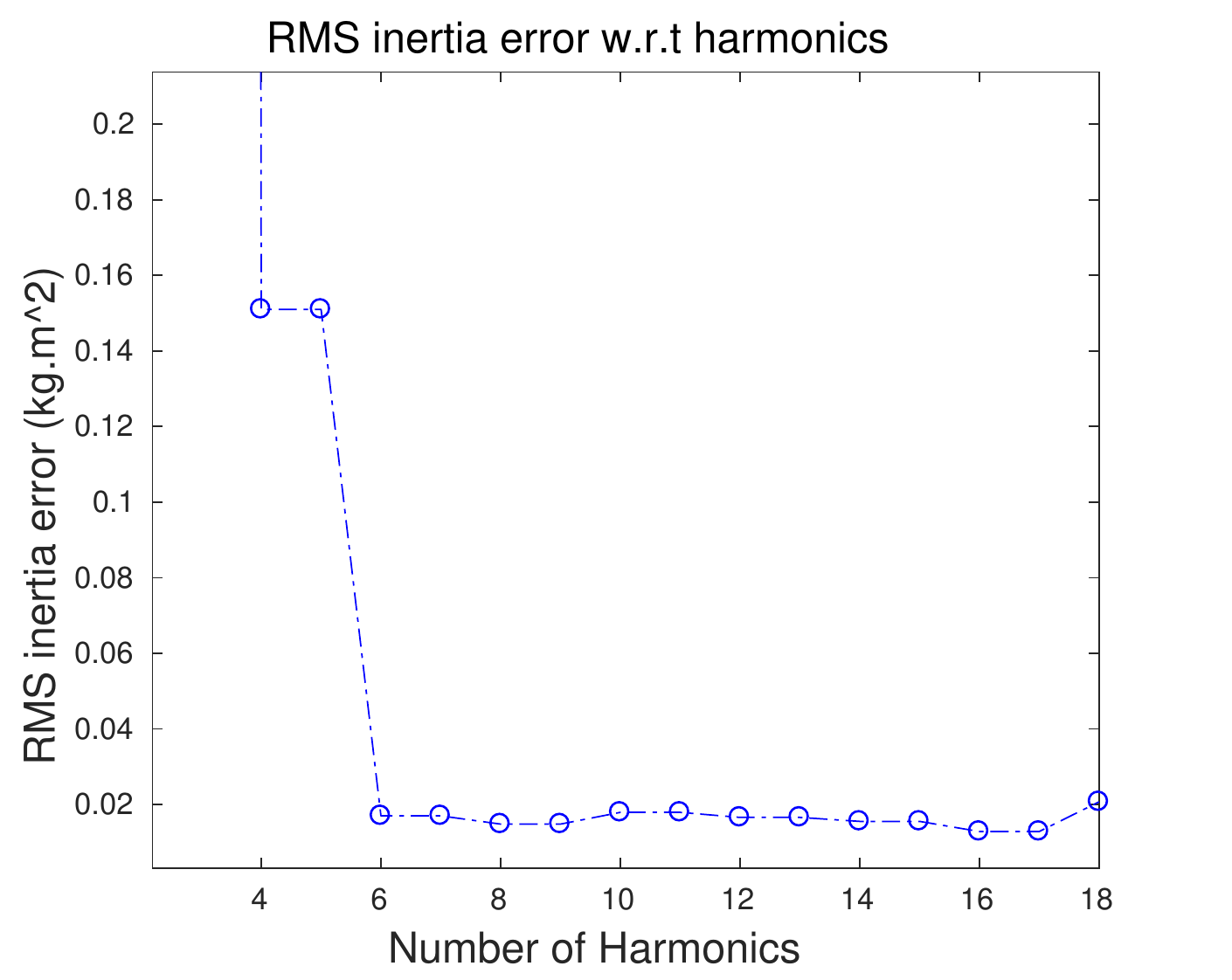}
	\subcaption{}\label{OE} 
	\endminipage%
	\minipage{0.333\linewidth}
	\includegraphics[width=\linewidth]{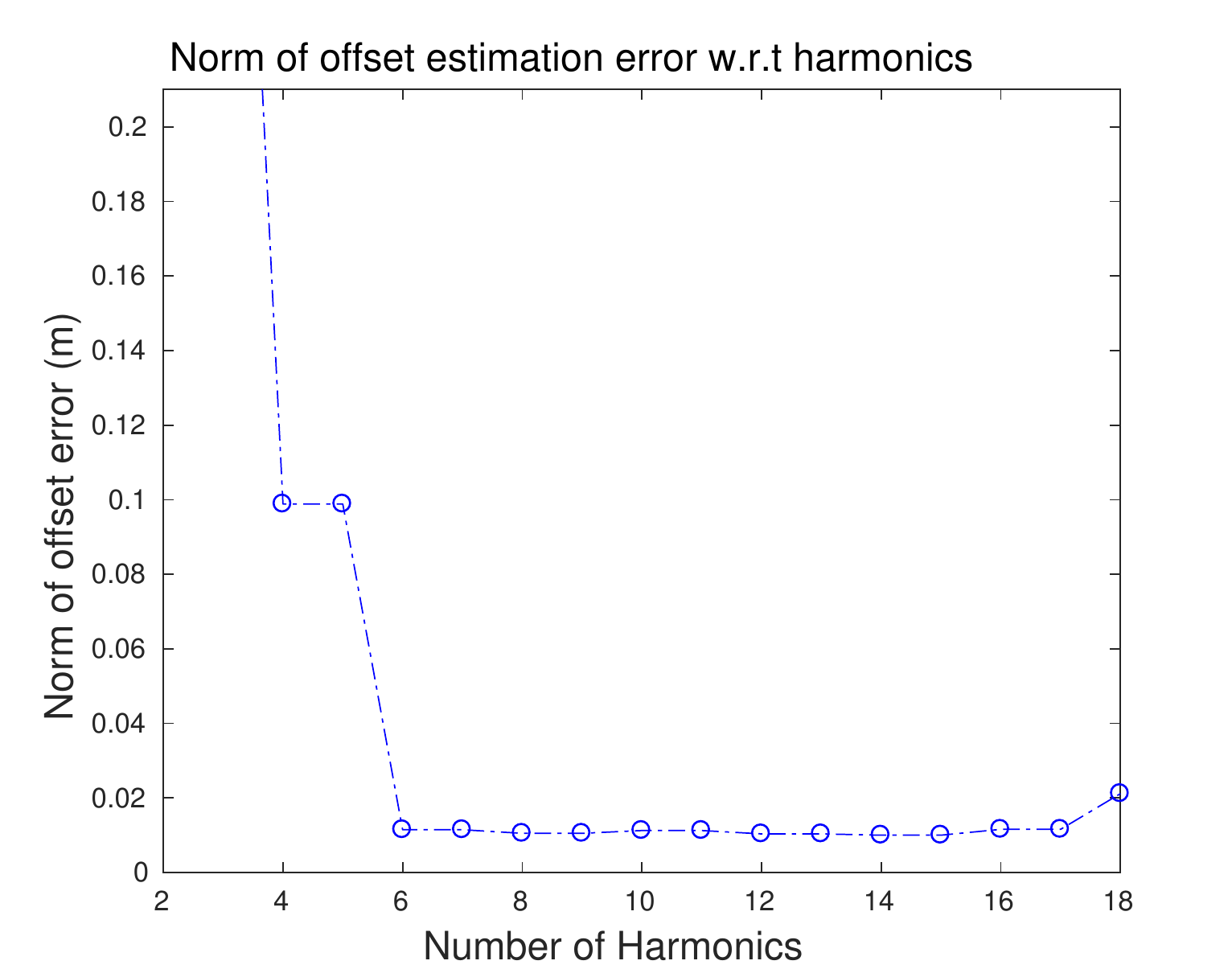}
	\subcaption{}\label{IE} 
	\endminipage%
	
	\setcounter{figure}{2}		
\end{figure*}
\begin{figure*}
	\captionsetup{justification=centering}\label{h}
	\caption{Variation in the accuracy of estimated parameters when a larger number of harmonics are used to filter and estimate the measured data. Plots show errors between actual parameter values and their estimates: a) Mass error, b) RMS error in inertia and c) norm of error in offset}
	\vspace{-1.5em}	
\end{figure*}
Trajectories for each of the 6 states, ($x, y, z, \phi, \theta, \psi$) from $X$ are finite Fourier series with \textit{n} harmonics given as:
\begin{equation} \label{eq4}
\begin{aligned}
X_i(t) &= a_{io} + \sum\limits_{k = 1}^n   \frac{a_{ik}}{\omega_fk}sin(\omega_fkt) - \frac{b_{ik}}{\omega_fk}cos(\omega_fkt)\\
\dot{X}_i(t) &= \sum\limits_{k = 1}^n   a_{ik}cos(\omega_fkt) + b_{ik}sin(\omega_fkt)\\
\ddot{X}_i(t) &= \sum\limits_{k = 1}^n    -a_{ik}\omega_fk sin(\omega_fkt) + b_{ik} \omega_fk cos(\omega_fkt)\\
\end{aligned}
\end{equation}
Where $\omega_f$ is a chosen angular frequency, $\omega_f= 2\pi / \mathbf{T}_f$, $\mathbf{T}_f$ being the period. The Fourier coefficients, $a_{io},a_{ik},b_{ik}$ for $k$ = 1 to $n$, and $i$ from 1 to 6 are denoted by vector $\pmb{\delta}$, which is obtained by optimizing one of the two criteria $J$:
\begin{enumerate}
	\item Minimizing the condition number of the $\mathbf{W}(\mathbf{X},\dot{\mathbf{X}},\ddot{\mathbf{X}})$ matrix which has been normalized, $J_1 = cond(\Sigma^{-0.5}\mathbf{W})$, where $\Sigma$ is the co-variance of the measured actuation values that lead to the calculation of applied forces and torques. With a small condition number, the estimation is less sensitive to noise from the measurements.
	\item Maximizing determinant of the Fisher information matrix, $J_2 = -log\vert \mathbf{W}^T\Sigma^{-1} \mathbf{W}\vert$, which is a measure of the amount of information the measured variable gives about the variable to be estimated.
\end{enumerate}

Using ${X}(t),\dot{{X}}(t),\ddot{{X}}(t)$ from (\ref{eq4}) for $t = 1$ to $N$, the optimization problem is formulated as:
\begin{align} & \displaystyle \min_{\delta} & & \displaystyle J(\mathbf{X},\dot{\mathbf{X}},\ddot{\mathbf{X}}) \\ \nonumber 
& \textrm{subject to:} & &   X(1) = 0 ,\dot{X}(1) = 0, \ddot{X}(1) = 0 \\ \nonumber & & & X(N) = 0 , \dot{X}(N) = 0,\ddot{X}(N) = 0 \\ \nonumber & & & X_{min} \leq X(t) \leq X_{max}\\
\nonumber & & & \dot{X}_{min} \leq \dot{X }(t) \leq \dot{X }_{max}
\\
\nonumber & & & \ddot{X}_{min} \leq \ddot{X }(t) \leq \ddot{X }_{max}
\end{align} 
such that the trajectory starts and ends with the robot at rest. The motion of the robot is constrained so that the generated trajectory does not have extreme values after optimization of the criteria. Note that this step does not require knowledge of the inertial parameters of the robot. The problem is non-convex, and the Multistart Solver from Matlab Global Optimization Toolbox\footnote{\url{https://www.mathworks.com/help/gads/}}  (MathWorks Inc.) was used for finding a solution.

\subsection{Executing the generated trajectory}
The system is made to track the excitation trajectory decribed in the previous section by using Non-linear Model Predictive Control (NMPC). This allows for actuation constraints to be included as a part of the control problem. We make an assumption  that initial values of the parameters of the free-flying robot are known. MPC makes use of those values in the system model for minimizing the error between desired and actual states with the least possible actuation over each horizon $t_h$ as:
\begin{align}\nonumber
& \underset{{X}(\cdot),u(\cdot)}{\text{minimize}}
& & \sum\limits_{{T} = t}^{t+t_h} ( \Vert e({X}_{des}({T}),{X}({T})) \Vert_Q^2 \; + \Vert u(\mathcal{T}) \Vert_P^2 )\\ 
& & &+\; \Vert e({X}_{des}(t+t_h),{X}(t + t_h))\Vert^2_{Q_N}\\\nonumber	
& \text{subject to:} 
& &\dot{{X}} = f({X}(t), u(t)) & \\\nonumber
& & &  u_{min} \leq u(t) \leq u_{max}
\end{align}

The function $e({X}_{des}(T), {X}(T))$ denotes the error between the states ${X}$. To avoid the issue of singularities in Euler Angle representations, the attitude-tracking controller is based on quaternions. The error function developed by Lee et. al \cite{c4} is used for calculating attitude errors. The non-linear model of system dynamics presented in section \ref{dyn} is represented by $\dot{\mathcal{X}} = f(\mathcal{X}(t), u(t))$, where $\mathcal{X} = [X,\dot{X}]$.$~Q,Q_N$ and $P$ are the weights given to each error. ACADO toolkit \cite{c8} was used to implement this controller. Note that the  controller makes use of initially known values of the inertial parameters, which do not correspond to the actual ones. Due to this, the obtained trajectory deviates from the actual one. Fig.\ref{fig:2} illustrates the excitation trajectory, and the resulting response when the robot tracks it using NMPC in simulation.

\section{PROCESSING OF MEASURED DATA}\label{sec:5}
Data measured at the time of tracking the excitation trajectory is to be used in parameter estimation, as shown in (\ref{eq14}) - (\ref{eq16}). The vector $\pmb{\tau}$ of applied forces and torques is calculated from the actuation inputs by using (\ref{eq01}). Estimates of position and angular orientation from the EKF would need to be differentiated in order to obtain angular velocities and linear and angular accelerations needed for constructing the $\pmb{\gamma}$ matrix. These measurements contain process and measurement noise, and differentiating them would mean introducing more noise in the calculations. In this section we present how the periodic and band-limited nature of the excitation trajectory is used to process measured data which contains noise.

\subsection{Noise filtering}\label{sec:NF}
An approach similar to that in \cite{c1} and \cite{c7} has been used here.
First, the robot is made to track $C$ cycles ($C$ is taken as 10 in our simulations) of the excitation trajectory. As the trajectory is periodic, the values of positions and attitude can be averaged over the time domain, in order to improve the signal-to-noise ratio of the data. \\
Second, as the excitation is band limited, we know that the output trajectory will contain at least the frequencies that were specified for the excitation trajectory. This way, the measured data can be converted to the frequency domain using Fast Fourier Transform (FFT) and then the expected frequencies can be retained while discarding the others as high frequency noise and setting their amplitude and phase to zero. The signal can then be converted back to the time domain. Further, the approach described in section \ref{ssec:Re} can be used to fit a Fourier series to the obtained signal, such that the derivatives can be calculated using the Fourier series, instead of numerical differentiation.\\
    However, other works using this approach have controllers with limited bandwidth on their manipulators to track the excitation trajectory. In our case, not only are the parameters of the current system different than those used by the MPC, but also the actuation input is subject to saturation. Consequently, the resulting trajectory might have more frequencies than the actual one, in addition to the high frequency measurement noise. The novelty in the proposed approach is, therefore, to use more harmonics for filtering and estimating the measured data than were used in the formulation of the excitation trajectory. There is a caveat here - using a large number of harmonics means that more noise could be carried forward, and including less frequencies means that information about the obtained trajectory is lost. Both cases will affect the accuracy of parameter estimation. This is illustrated by Figs. \ref{ME}, \ref{OE} and \ref{IE}, where errors in the estimated parameters with respect to their actual values are plotted. It is also seen that the same value of $n$ may not give the least error for all parameters to be estimated. Therefore, a method to pick the value of n is needed, which gives the least estimation error for as many parameters as possible, without knowing their ground truth values.
    Such an approach for selecting the number of harmonics is proposed in section V-B.

 \begin{figure}
 	\centering
 	\includegraphics[width=80mm,scale=0.5]{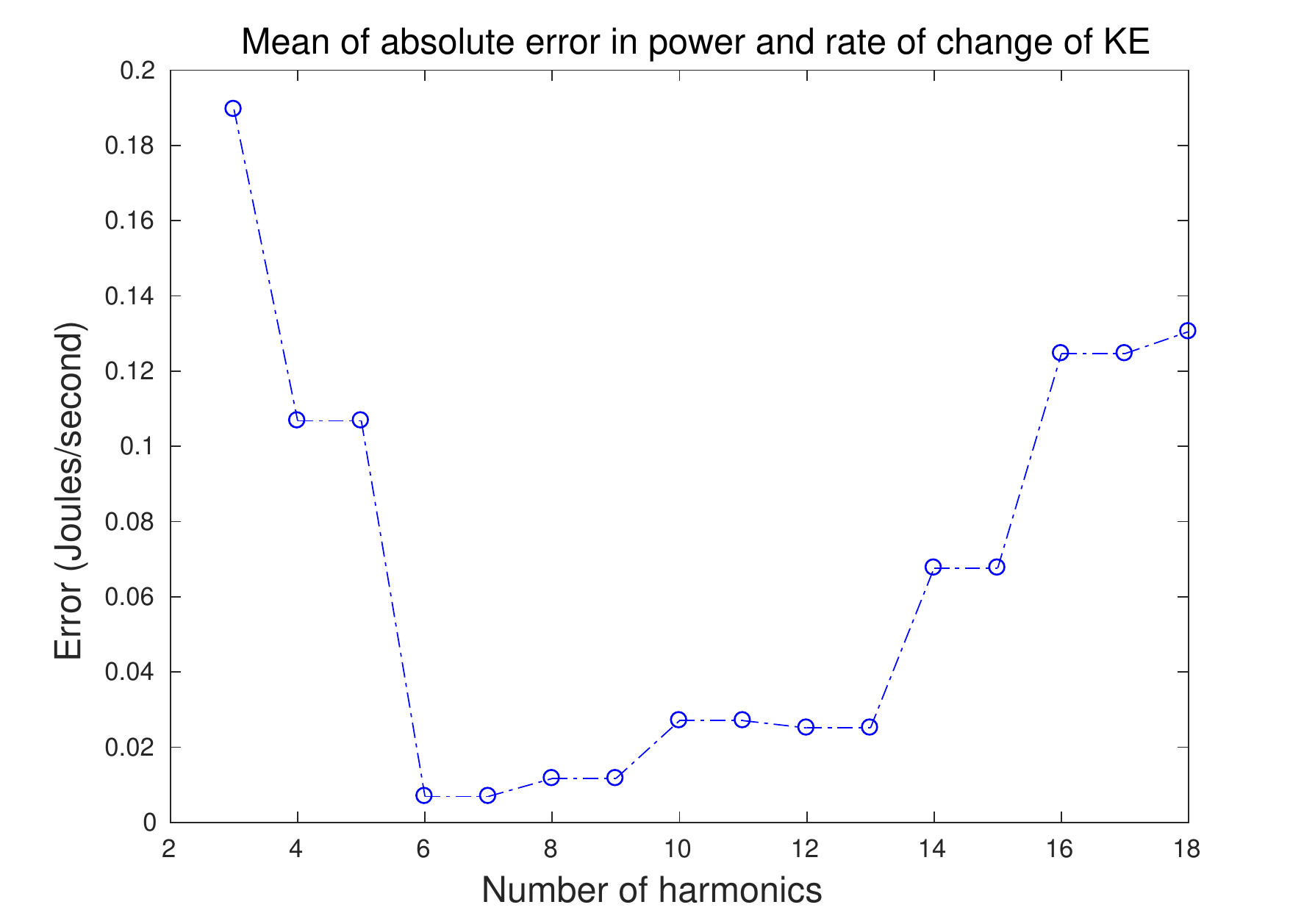}
 	\captionsetup{justification=centering}
 	\caption{Number of harmonics that give the least error between input power and the rate of kinetic energy}\label{EH}
 	  	\vspace{-1.0em}	
 \end{figure}

\subsection{Selecting the number of harmonics}\label{H}
Equations for the total kinetic energy of the system with new inertial parameters is given as 
\begin{align} \label{KE}
T &= \frac{1}{2}m\dot{\textbf{\textit{p}}}_s^T\dot{\textbf{\textit{p}}}_s +  \frac{1}{2}\pmb{\omega}^T\mathbf{J}_s\pmb{\omega} 
\end{align}
where the first time gives the translation kinetic energy component, $T_{trans}$ and the second, the rotational kinetic energy component or $T_{rot}$.
The kinetic energy rate is obtained by differentiating (\ref{KE}), to give:
\begin{align}\label{dKE}
\dot{T} &=  m\ddot{\textbf{\textit{p}}}_s^T
\dot{\textbf{\textit{p}}}_s + \pmb{\omega}^{T}\left(\mathbf{J}_s\dot{\pmb{\omega}} +  \pmb{\omega} \times \mathbf{J}_s\pmb{\omega}\right)
\end{align}
Using (\ref{eq00}), (\ref{eq1}) and (\ref{eq2}), the above expression can be written for power in the body frame as:
\begin{align}\label{Power}
\dot{T} = \mathbf{F}^T(\mathbf{R}^{-1}\dot{\textbf{\textit{p}}}_c + \pmb{\omega} \times {\textit{\textbf{\textit{p}}}}_{off} ) + (M -\textbf{\textit{\textbf{\textit{p}}}}_{off}\times\mathbf{F})^T\pmb{\omega}
\end{align}
Re-arranging (\ref{dKE}) and (\ref{Power}) after expressing all quantities in the body frame, we obtain
\begin{equation}
\begin{aligned}\label{Fin1}
&\mathbf{F}^T\mathbf{R}^{-1}\dot{\textbf{\textit{p}}}_c  + M^T\pmb{\omega} - \pmb{\omega}^T( \textbf{\textit{\textbf{\textit{p}}}}_{off}\times\mathbf{F}) + \mathbf{F}^T( \pmb{\omega} \times {\textit{\textbf{\textit{p}}}}_{off} ) = \\& m(\mathbf{R}^{-1}\ddot{\textbf{\textit{p}}}_s) ^T
\mathbf{R}^{-1}\dot{\textbf{\textit{p}}}_s  + \pmb{\omega}^{T}\left(\mathbf{J}_s\dot{\pmb{\omega}} +  \pmb{\omega} \times \mathbf{J}_s\pmb{\omega}\right)
\end{aligned}
\end{equation}
Using the properties of triple product, (\ref{Fin1}) is reduced to
\begin{equation}\label{Fin0}
\resizebox{1\hsize}{!}{$\mathbf{F}^T\mathbf{R}^{-1}\dot{\textbf{\textit{p}}}_c  + M ^T\pmb{\omega} = m(\mathbf{R}^{-1}\ddot{\textbf{\textit{p}}}_s) ^T
	\mathbf{R}^{-1}\dot{\textbf{\textit{p}}}_s  + \pmb{\omega}^{T}\left(\mathbf{J}_s\dot{\pmb{\omega}} +  \pmb{\omega} \times \mathbf{J}_s\pmb{\omega}\right)$}
\end{equation}
where (\ref{eq00}) and the component in curly brackets of (\ref{eq1}) give the expansion of $\dot{\textbf{\textit{p}}}_s$ and $\ddot{\textbf{\textit{p}}}_c$ in terms of the measured velocities and accelerations of the vehicle centre of mass $\textbf{\textit{p}}_s$ and the center of mass offset, $\textbf{\textit{p}}_{off}$.
\begin{figure*}[h!]
	
	\centering
	\minipage{0.333\linewidth}
	\includegraphics[width=\linewidth]{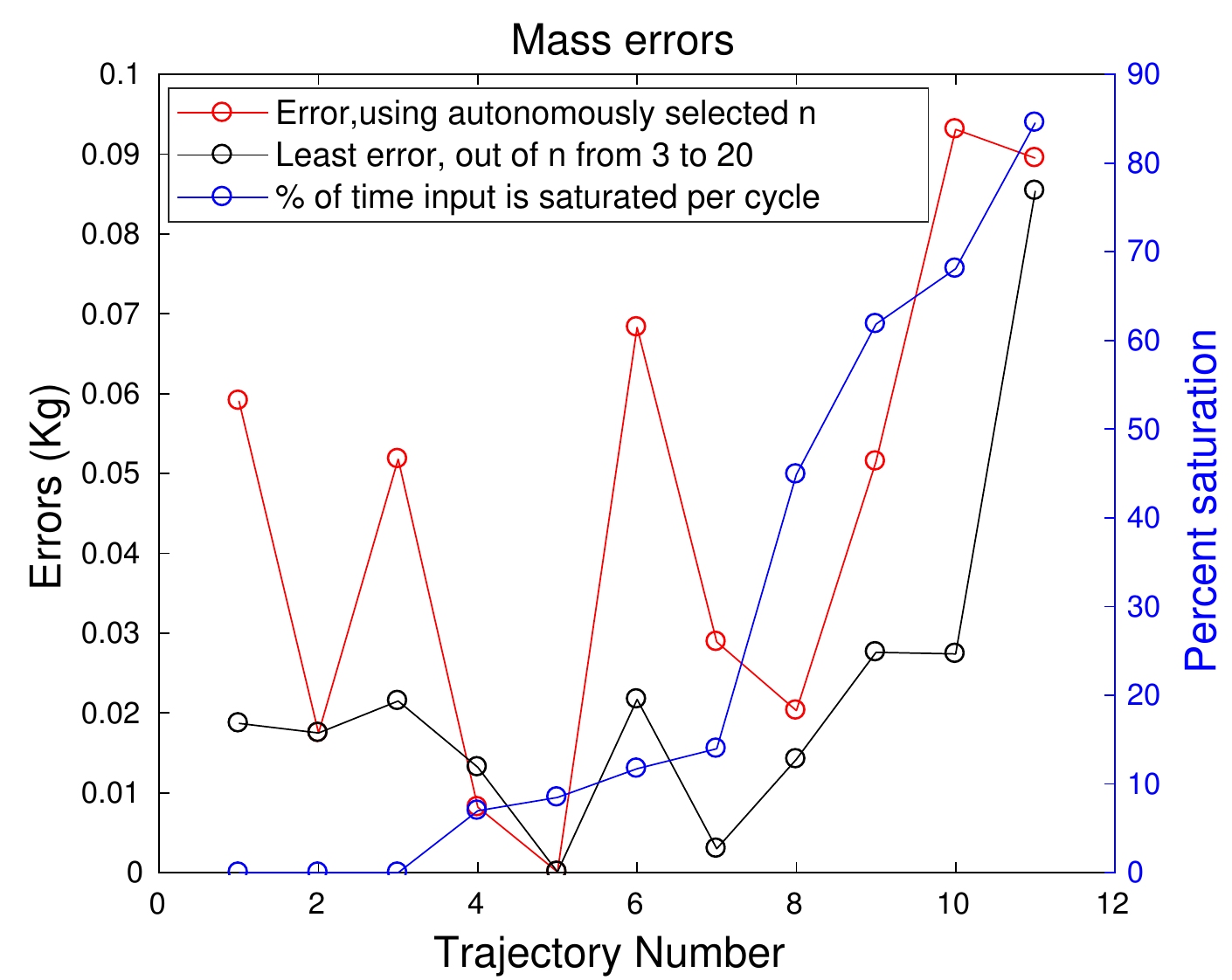}
	\subcaption{}\label{Mass_sat} 
	\endminipage%
	\minipage{0.333\linewidth}
	\includegraphics[width=\linewidth]{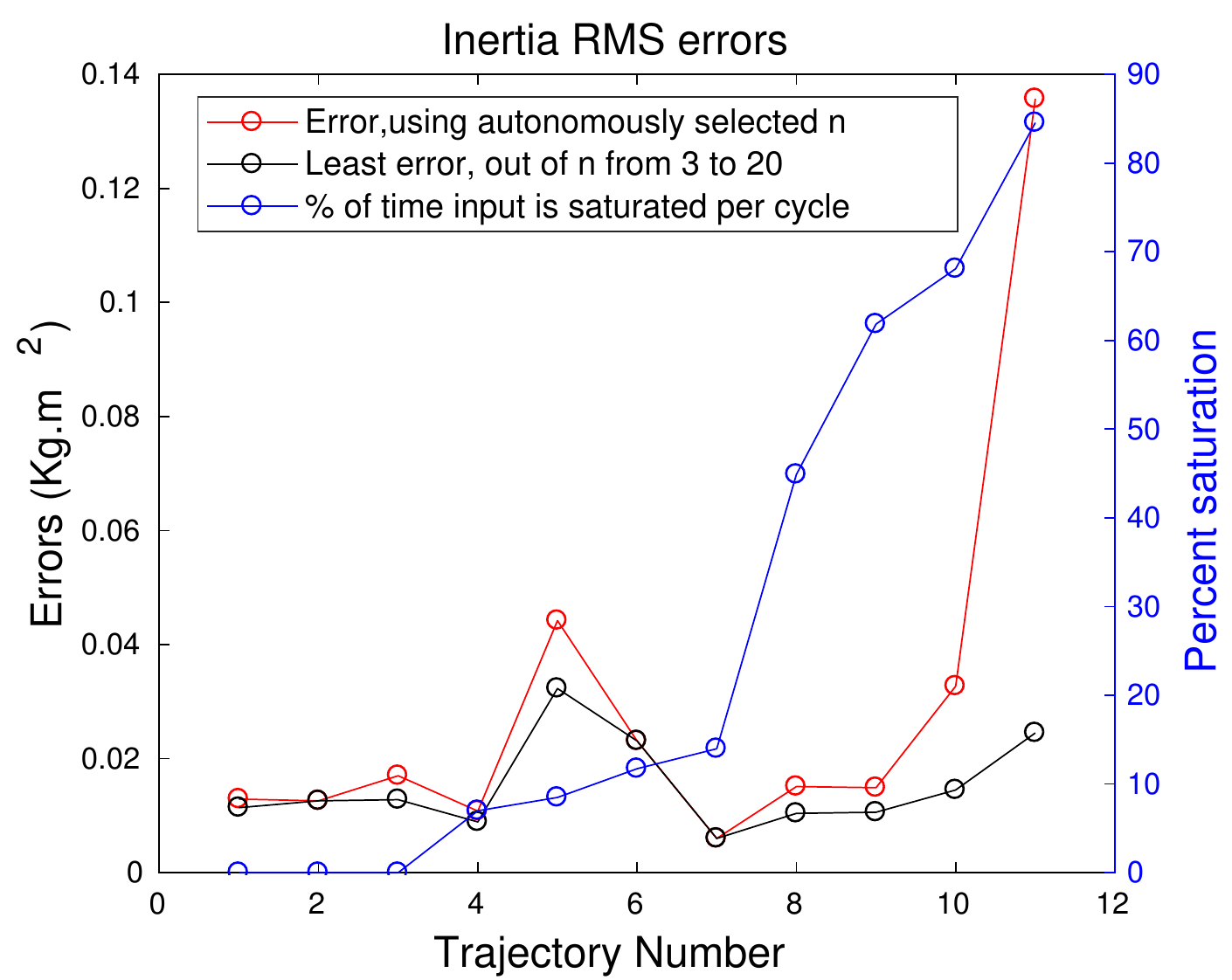}
	\subcaption{}\label{Off_sat} 
	\endminipage%
	\minipage{0.333\linewidth}
	\includegraphics[width=\linewidth]{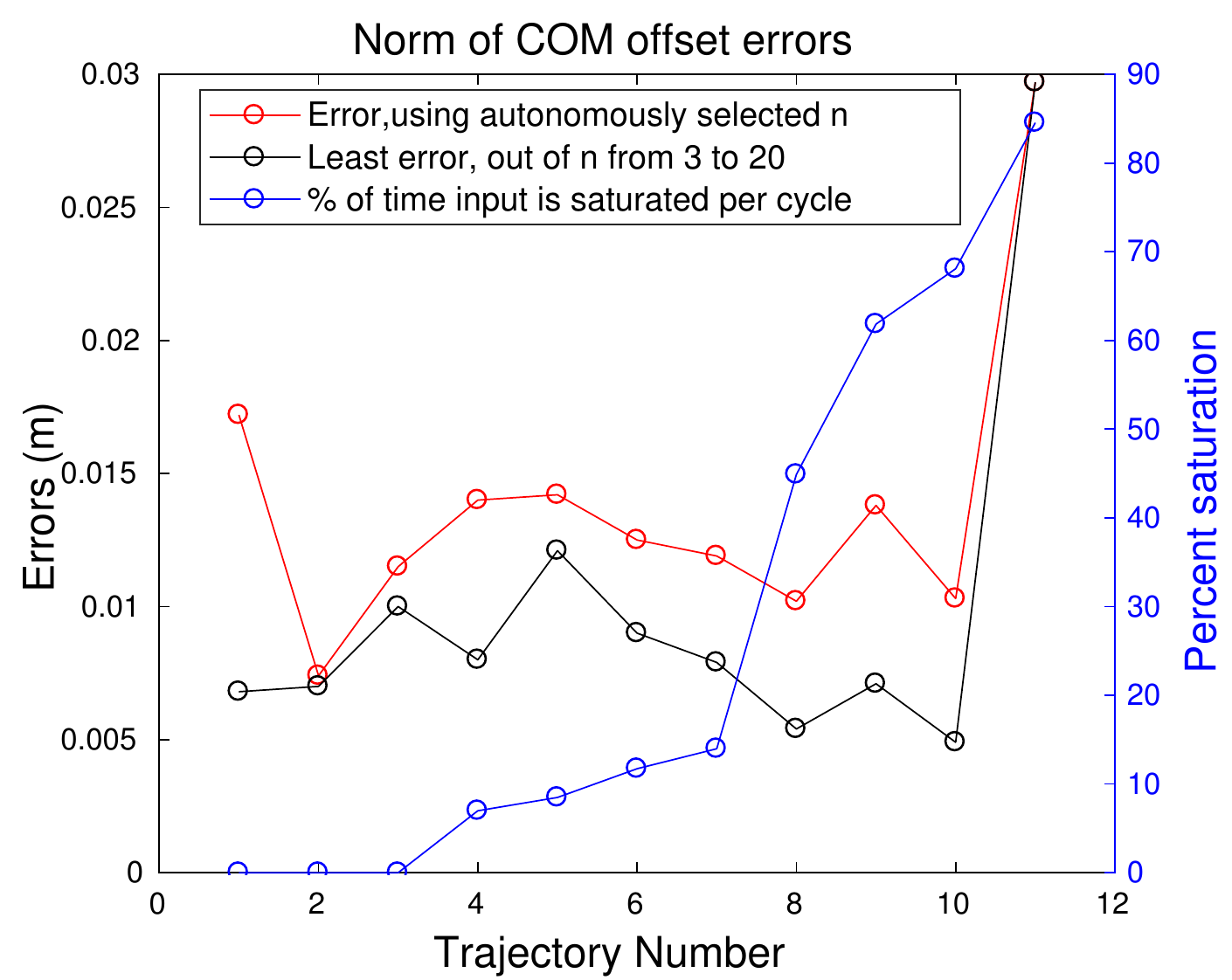}
	\subcaption{}\label{I_sat} 
	\endminipage%
			\vspace{-1.5em}	
				\setcounter{figure}{4}	
\end{figure*}

\begin{figure*}
	\captionsetup{justification=centering}\label{H1}
	\caption{Parameter estimation errors when (i) $n^*$ is selected autonomously and (ii) all harmonics, $n$ from 3 to 20 are tested and the least errors are plotted for a) mass, b) inertia and c)com offset parameters. The excitation trajectories are arranged in increasing order of percent input saturation (saturation observed on one or more of the inputs) encountered per cycle while tracking them, which is also plotted. }
	\vspace{-2.0em}
\end{figure*}

Therefore, the components on the right give the input power whereas those on the left give the rate of kinetic energy. These will be referred to as $P$ and $\dot{T}$, respectively. $P$ makes use of the input forces and torques, and measured velocities, while  the calculation of $\dot{T}$ needs estimates of the inertial parameters and the measured velocities and accelerations.
As mentioned in section \ref{sec:NF}, for an ideal number of harmonics $n$, the measured data will have the least amount of high frequency noise and give the most accurate parameter estimation. Therefore, we expect that the error between $P$ and $\dot{T}$ will be the least with this $n$. The method used for harmonic selection is the following: For all number of harmonics between 3 and 15, steps detailed in sections (A) and (B) are carried out, and the estimated parameters corresponding to the $n$ for which the mean value of $ \vert P - \dot{T} \vert$ is the least, is chosen. Fig. \ref{EH} shows an example of how this error varies as more harmonics are used, for the case shown in Figs.\ref{ME}, \ref{OE} and \ref{IE}. Here, 7 harmonics give the least error, so $n$ is chosen as 7.

 \subsection{Estimating Fourier Coefficients}\label{ssec:Re}
 Although high frequency noise from the measured data is removed, noise corresponding to the retained frequencies cannot be eliminated. So, with the selected number of harmonics $n$, we fit a truncated Fourier series to the obtained data as:  
 \begin{align}\nonumber
 & \underset{\hat{\delta}}{\text{minimize}}
 & & \sum\limits_{{T} = 0}^{\mathbf{T}_f}  \Vert {X}_{measured}({T})-\hat{X}({T}) \Vert ^2\\ 
 \end{align}
 where $\hat{\mathbf{X}}$ is related to the estimated vector of Fourier series co-efficients $\hat{\pmb{\delta}}$ by (\ref{eq4}). However, as it is known that the least squares fit of a truncated Fourier series is equivalent to finding its Fast Fourier Transform, FFT can be used instead of a least square problem. $\dot{\hat{X}}(t),\ddot{\hat{X}}(t)$ are then found analytically by using the estimated coefficients, in order to construct the $\pmb{\gamma}$ matrix. The least square solution in (\ref{eq16}) is then used to calculate estimates of the inertial parameters.

\section{SIMULATION RESULTS}\label{sec:6}
This section presents the results of the discussed parameter estimation method. Simulations are carried out for the Space CoBot \cite{c5}, considering a scenario where it has grasped an object and the new inertial parameters of the system are unknown. Eleven excitation trajectories were generated, each of 10 seconds, by minimizing either criteria $J_1$ and $J_2$. These trajectories were Fourier Series with 3 harmonics for each of the 6 states in $\mathbf{X}$. The robot was made to track these trajectories for 10 cycles (i.e.,100 seconds). The mass of the Space CoBot is 6.047 Kg. Its inertia tensor is taken to be diagonal, with components $J_{xx} = 0.0453$, $J_{yy} = 0.0417 $ and $J_{zz} = 0.0519 $. Two unmodeled payloads, load 1 and 2, of 1.2 and 0.5 Kg, are considered.      
\subsection{Testing the method of selection of number of harmonics}\label{ssec:VI}
In order to test the harmonics selection method, the number of harmonics, $n$, used for filtering and estimation of the measured data was iteratively increased from 3 to 20. Then: \\ (1) For our method, the difference between the input power and rate of change of kinetic energy was calculated each time, and the value of $n^*$ at which the \textit{least $\vert P - \dot{T}|$ error} was found, was chosen. The parameter estimates corresponding to this value of n is the final result or $\hat{\pi}_n^*$. $\hat{\pi}_n^*$ was then compared to the actual values $\pi$ for calculating the estimation errors. (2) In contrast, the least estimation errors, or \textit{least $ err(\hat{\pi}_{n,param},\pi_{param})$ for all n from 3 to 20} were found for the parameters ($param$) of mass, centre of mass offset, and inertia.
 The results of (1) and (2) for 11 excitation trajectories are plotted in Figs. \ref{Mass_sat}, \ref{I_sat} and \ref{Off_sat}. A payload of 1.2 Kg was considered in both cases. The trajectories are arranged in increasing order of saturation, which is also plotted. Overall, it is observed that the estimation errors from case (1) follow the same trend as the errors from case (2). Though this method of harmonic selection can cope with low amounts of saturation, high amount of saturation does not seem to be favourable and the estimation errors are very high for all 3 parameters.
 \subsection{Comparing the performance of the criteria}\label{ssec:V}
  The plots show that a high percentage of input saturation seems to affect the accuracy of parameter estimation to some extent. Further, the amount of saturation that will occur cannot be predicted before trajectory execution. So, excitation trajectories where input saturation does not occur were used to compare results obtained with the two criteria, for load 1 and load 2. Out of the 11 exciting trajectories considered, 5 optimized criteria $J_1$ and 5, $J_2$. The proposed parameter estimation approach was followed completely, with autonomous selection of harmonics, and the resulting estimation errors are plotted in Fig.7. The figures show that in general, higher mass estimation errors are obtained with criteria $J_2$ while criteria $J_1$ gives higher inertia and COM offset errors.

\begin{figure*}[h!]
	
	\centering
	\minipage{0.333\linewidth}
	\includegraphics[width=\linewidth]{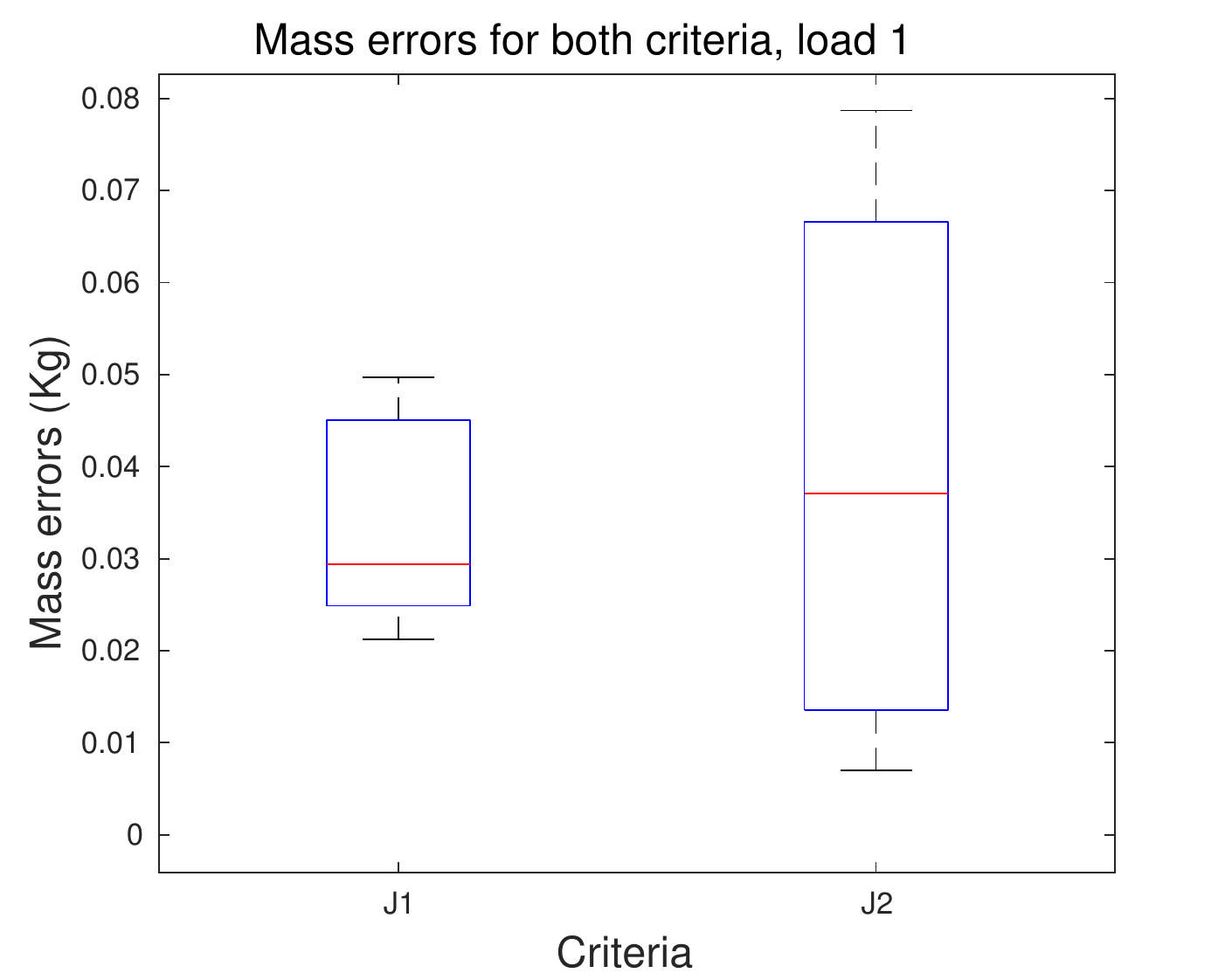}
	\subcaption{}\label{M1} 
	\endminipage%
	\minipage{0.333\linewidth}
	\includegraphics[width=\linewidth]{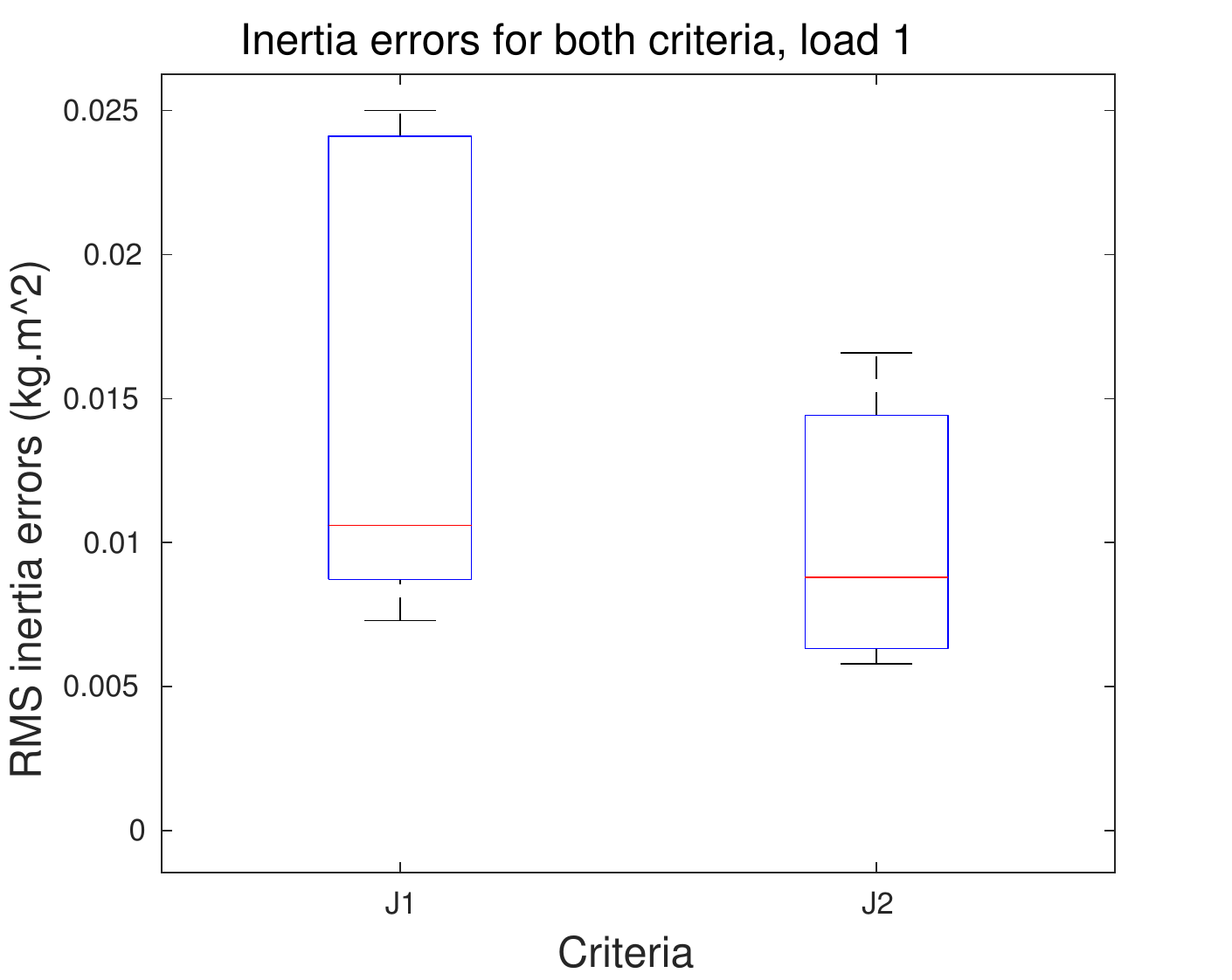}
	\subcaption{}\label{O1} 
	\endminipage%
	\minipage{0.333\linewidth}
	\includegraphics[width=\linewidth]{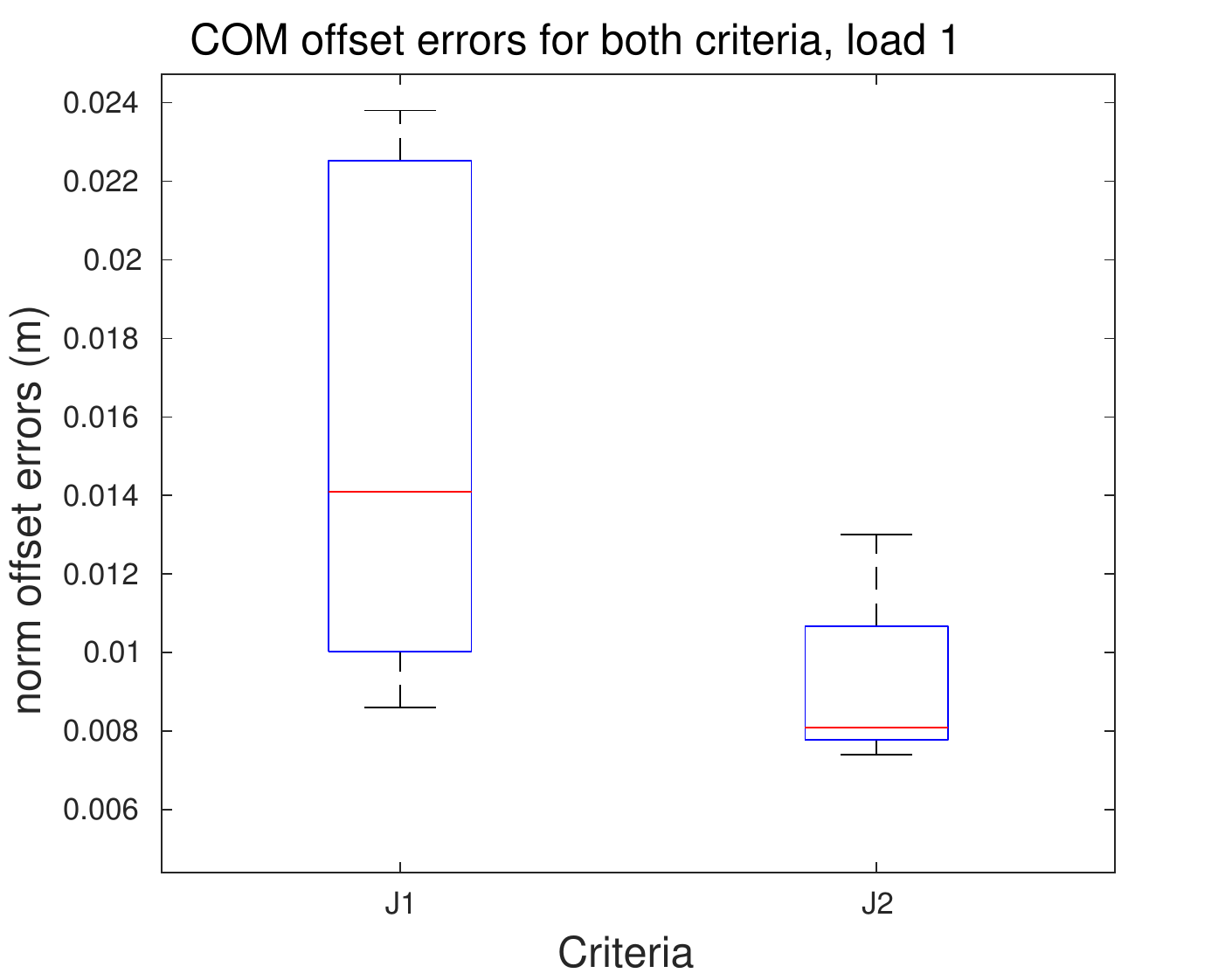}
	\subcaption{}\label{I1} 
	\endminipage%
	\centering\\
	\minipage{0.333\linewidth}
	\includegraphics[width=\linewidth]{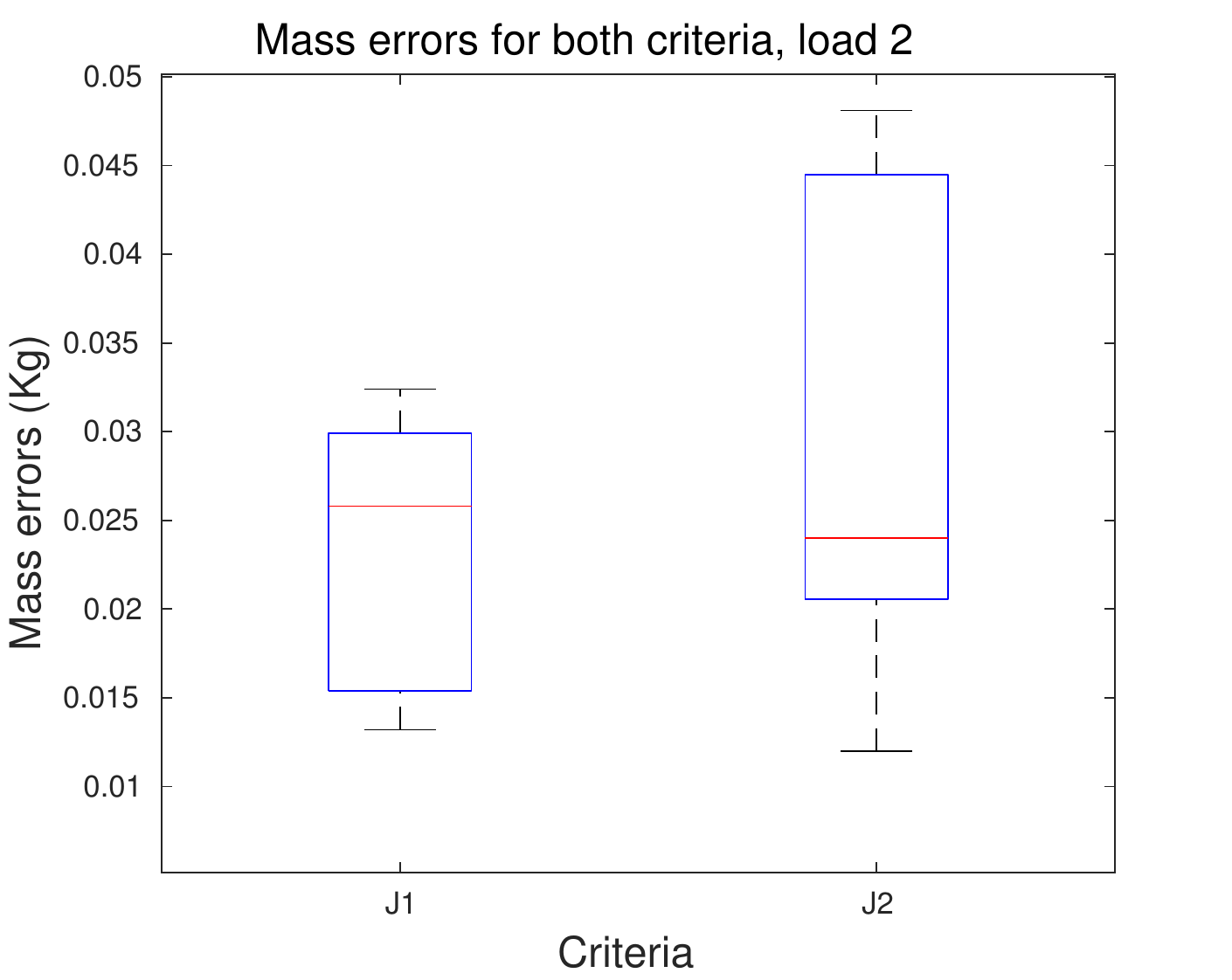}
	\subcaption{}\label{M2} 
	\endminipage%
	\minipage{0.333\linewidth}
	\includegraphics[width=\linewidth]{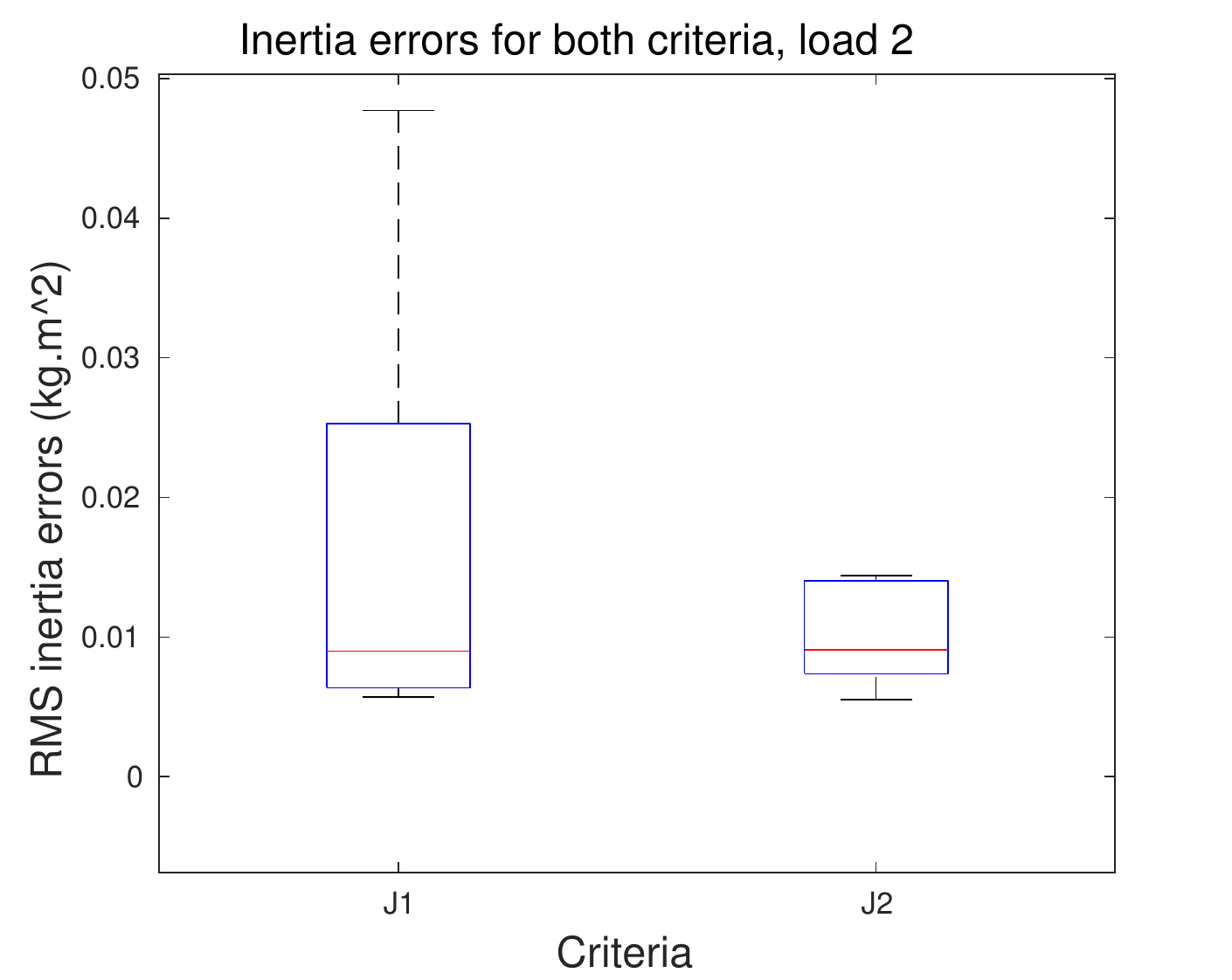}
	\subcaption{}\label{O2} 
	\endminipage%
	\minipage{0.333\linewidth}
	\includegraphics[width=\linewidth]{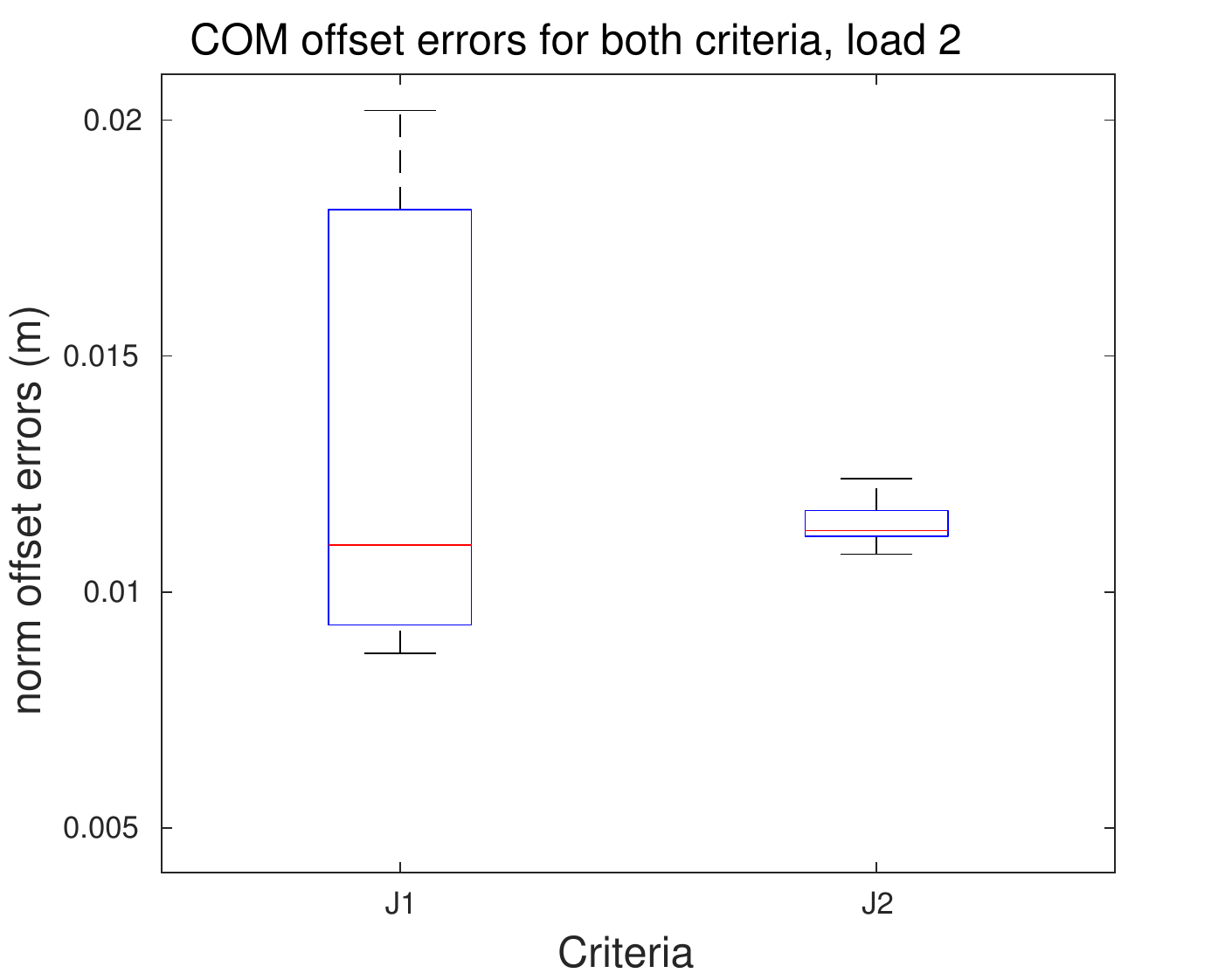}
	\subcaption{}\label{I2} 
	\endminipage%
	\vspace{-1.5em}	
	\setcounter{figure}{5}	
\end{figure*}
\begin{figure*}[h!]
	\captionsetup{justification=centering}
	\caption{Parameter estimation errors for mass, inertia and centre of mass parameters for 2 unmodeled loads obtained with 10 exciting trajectories, half of them found by optimizing $J_1$ and the other half by $J_2$  }
	\vspace{-1.7em}
\end{figure*}

\section{Conclusion}\label{Conclusion}
This paper proposed a method for estimating inertial parameters of a free-flying robot after it has grasped an object. Fourier-series based excitation trajectories were tracked with Non-linear Model Predictive Control, and truncated Fourier series were fit to the measured data in order to filter noise and estimate the obtained trajectory. The tracked trajectory was shown to have more harmonics than the desired one, which created the need of a method that would find the best number of harmonics that fits the measured data, but not the high frequency noise contained in it. Thus, an energy balance based approach was introduced for this purpose. The effect of input saturation on this method was studied. Finally, the results of parameter estimation using different exciting trajectories obtained by optimizing the two criteria were compared for 2 cases of unmodeled load. Simulation results show that in the presence of a lot of saturation, the performance of harmonic selection as well as parameter estimation is affected. At the time of the calculation of exciting trajectories, constraints are placed on the motion of the robot, but control input limits are not accounted for explicitly. Future work will focus on tackling the problem of input saturation. The dynamics of the manipulator will be included in the robot model, and the feasibility of inertial parameter estimation of the system and the arm, with excitation only in the joint space will be tested. Furthermore, this method will be integrated in grasping and transportation processes.

\end{document}